\documentclass[conference]{IEEEtran}

\usepackage{tabularx}
\usepackage{algorithm,algorithmic,multirow,hhline}

\usepackage{tabularx}
\usepackage{array}

\usepackage{hyperref}
\usepackage{url}
\usepackage[noadjust]{cite}
\usepackage{amsmath,graphicx}
\usepackage{placeins}
\usepackage{amsmath,epsfig}
\usepackage{epstopdf}
\usepackage{amssymb}

\usepackage{amssymb}
\usepackage{caption}
\usepackage{comment}
\usepackage{subfigure}
\usepackage{pstricks}
\usepackage{subfloat}
\usepackage{setspace}

\usepackage{enumitem}
\usepackage{booktabs}       
\usepackage{stfloats} 

\def\lamb{\boldsymbol{\lambda}}  
\def\sm{ \small }
\def\nm{ \normalsize} 
\def\vs{ \vspace{-10pt} }

\begin{document}

\title{ \fontsize{23}{28}\selectfont L-FGADMM: Layer-Wise Federated Group ADMM for Communication Efficient Decentralized Deep Learning} 

\author{\IEEEauthorblockN{Anis Elgabli, Jihong Park, Sabbir Ahmed, and Mehdi Bennis}
	\IEEEauthorblockA{\emph{University of Oulu, Finland} \\\{anis.elgabli, jihong.park, sabbir.ahmed, mehdi.bennis\}@oulu.fi \\
	}}

\maketitle

\begin{abstract}
This article proposes a communication-efficient decentralized deep learning algorithm, coined layer-wise federated group ADMM (L-FGADMM). To minimize an empirical risk, every worker in L-FGADMM periodically communicates with two neighbors, in which the periods are separately adjusted for different layers of its deep neural network. A constrained optimization problem for this setting is formulated and solved using the stochastic version of GADMM proposed in our prior work. Numerical evaluations show that by less frequently exchanging the largest layer, L-FGADMM can significantly reduce the communication cost, without compromising the convergence speed. Surprisingly, despite less exchanged information and decentralized operations, intermittently skipping the largest layer consensus in L-FGADMM creates a regularizing effect, thereby achieving the test accuracy as high as federated learning (FL), a baseline method with the entire layer consensus by the aid of a central entity.

\end{abstract}


\section{Introduction}
\label{sec:intro}
Interest in data-driven machine learning (ML) is on the rise, but difficulties in securing data still remain~\cite{park2018wireless,Lu:19}. Mission critical applications aggravate this challenge, which require a large volume of up-to-date data for timely coping with local environments even under extreme events~\cite{Park20:xURLLC}. Mobile devices prevailing at the network edge are a major source of these data, but their user-generated raw data are often privacy-sensitive (e.g., medical records, location history, etc.). In view of this, distributed ML has attracted significant attention, whereby the parameters of each model, such as the weights of a neural network (NN), are exchanged without revealing raw data~\cite{park2018wireless,Lu:19,Park:2019:FLlet,Smith:FLSurvey}. However, with deep NN architectures, the communication payload sizes may be too large, and hinder the performance of distributed ML, spurring a quest for communication-efficient distributed ML solutions.



\begin{figure}[t]
\centering
\includegraphics[width=\columnwidth]{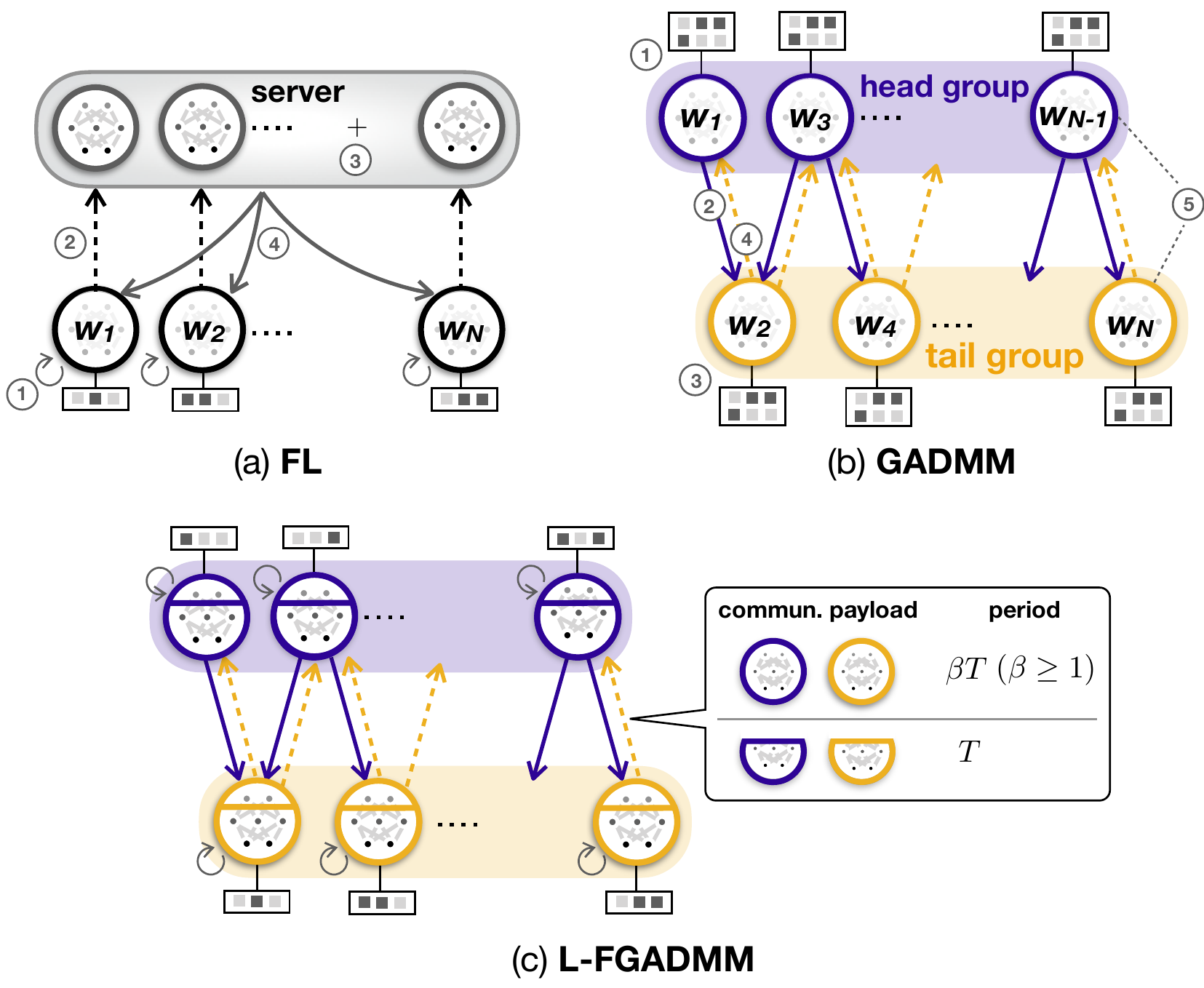} \vskip -5pt
\caption{\sm Operational structures of (a) federated learning (FL), (b) group ADMM (GADMM), and (c) our proposed \emph{layer-wise federated GADMM (L-FGADMM)} combining FL and GADMM, while less frequently exchanging the largest NN layer compared to the other~layers.} 
\label{fig1}
\end{figure}

\vspace{3pt}\noindent\textbf{Federated Learning (FL).}\quad
A notable communication-efficient distributed ML framework is FL~\cite{Brendan17,samarakoon2018federated,KimCL:19}. Each device, or worker, in FL stores its own dataset and an NN, and locally trains the NN via stochastic gradient descent (SGD). As shown in Fig.~\ref{fig1}a, the weight parameters of the local NN are uploaded to a parameter server at a regular interval. The server thereby produces the global average weight parameters that are downloaded by each worker. These FL operations are summarized by its server-aided \emph{centralized architecture}, \emph{random data sampling} per SGD iteration, and \emph{periodic communication} at an interval of multiple SGD iterations. However, the inherently centralized architecture of FL is ill-suited for mobile devices located faraway from the server. Due to the limited transmission power and energy, these devices may easily loose connectivity, calling for decentralized ML methods.




\vspace{3pt}\noindent\textbf{Group ADMM (GADMM).}\quad
To achieve fast convergence while minimizing the number of communication links under a \emph{decentralized architecture}, GADMM was proposed in our prior works~\cite{elgabli2019gadmm,anisgadmmgl,elgabli2019qgadmm}. Under GADMM, each worker communicates only with its two neighboring workers. To this end, as illustrated in Fig.~\ref{fig1}b, GADMM first divides the workers into head and tail groups. The workers in the same group update their model parameters in parallel, while the workers in different groups update their models in an alternating way, after communicating with neighbors in different groups. Nonetheless, the effectiveness of GADMM was shown only for convex loss functions without exploring deep NN architectures. What's more, GADMM assumes that every worker identically stores the \emph{full batch data} and runs gradient descent (GD) with \emph{immediate communication} per GD iteration. These assumptions are ill-suited for the user-generated nature of data and communication efficiency, motivating us to seek for a federated and decentralized solution.


\vspace{3pt}\noindent\textbf{Layer-wise Federated GADMM (L-FGADMM).}\quad
To bridge the gap between FL and GADMM, in this article we propose L-FGADMM, by integrating the periodic communication and random data sampling properties of FL into GADMM under a deep NN architecture. To further improve communication efficiency, as illustrated in Fig.~\ref{fig1}c, L-GADMM applies a \emph{different communication period to each layer}. By exchanging the largest layer $2$x less frequently than the other layers, our results show that L-FGADMM achieves the same test accuracy while saving $48.8$\% and $60.8$\% average communication cost, compared to the case using the same communication period for all layers and FL,~respectively.







\vspace{5pt}\noindent\textbf{Related Works.}\quad
Towards improving communication efficiency of distributed ML, under centralized ML, the number of communication rounds can be reduced by collaboratively adjusting the training momentum~\cite{Liu:2019aa,Yu:2019aa}. On the other hand, the number of communication links can be decreased by collecting model updates until a time deadline~\cite{Wang:2019aa}, upon the values sufficiently changed from the preceding updates~\cite{chen2018lag,sun2019communication}, or based on channel conditions~\cite{FL_Nishio,YangQuekPoor:2019aa,Chen:20019aa}. Furthermore, the communication payload can be compressed by 1-bit gradient quantization \cite{Bernstein:2018aa}, multi-bit gradient quantization~\cite{sun2019communication}, or weight quantization with random rotation \cite{pap:jakub16}. Alternatively, instead of model parameters, model outputs can be exchanged for large models via knowledge distillation~\cite{Jeong18,Ahn:2019aa}. Similar principles are applicable for communication-efficient decentralized ML. Without any central entity, communication payload sizes can be reduced by a quantized weight gossiping algorithm \cite{Koloskova:2019aa}, ignoring communication link reduction. Alternatively, the number of communication links and rounds can be decreased using GADMM proposed in our prior work~\cite{elgabli2019gadmm}. Furthermore, by integrating stochastic quantization into GADMM, quantized GADMM (Q-GADMM) was proposed to reduce communication rounds, links, and payload sizes altogether~\cite{elgabli2019qgadmm}. To achieve the same goals, instead of quantization as in Q-GADMM, L-FGADMM applies a layer-wise federation to GADMM under deep NN architectures. Combining both quantization and layer-wise federation is deferred to future work.



\section{Problem Formulation}
\label{probForm}
We consider $N$ workers, each of which stores its own batch of input samples and runs a deep NN model comprising $L$ layers. Hereafter, the subscript $n$ identifies the workers, and the superscript $\ell$ indicates the layers. The $n$-th worker's model parameters (i.e., weights and biases) are denoted as $\boldsymbol{\phi}_n=\{\boldsymbol{\theta}_n^1, \boldsymbol{\theta}_n^1,\cdots,\boldsymbol{\theta}_n^L\}$, where $\boldsymbol{\theta}_n^l \in\mathbb{R}^d$ is a $d$-dimensional vector whose elements are the model parameters of the $\ell$-th layer.

Every worker has its local loss function $f_n(\boldsymbol{\phi}_n)$, and optimizes its local model $\boldsymbol{\phi}_n$ such that the global average loss $\sum_{n=1}^N f_n(\boldsymbol{\phi}_n)$ can be minimized, at which all local models reach a consensus on a global model $\boldsymbol{\phi}=\boldsymbol{\phi}_1=\cdots=\boldsymbol{\phi}_N$. To solve this problem in parallel, each worker runs a first-order iterative algorithm by selecting a mini-batch $X_n^{(k)}$ at the $k$-th iteration, while communicating with other workers to ensure the consensus across $\{\boldsymbol{\phi}_n\}$. Unfortunately, this consensus requires exchanging local models, incurring huge communication overhead for deep NNs. To reduce the communication payload sizes, we instead consider a consensus across $\{\boldsymbol{\theta}_n^\ell\}$, leading to the following problem formulation:
\begin{align}
    \text{Minimize}\quad &\frac{1}{K N} \sum_{k=1}^K\sum_{n=1}^N f_n(\boldsymbol{\phi}_n, X_n^{(k)})
       \label{com_admm}\\
       \text{subject to}\quad  &
    \boldsymbol{\theta}_n^\ell=\boldsymbol{\theta}_{n+1}^\ell.
    \label{com_admm_c2}
\end{align}
The constraint \eqref{com_admm_c2} implies a per-layer consensus between the $n$-th and $(n+1)$-th workers. This enables layer-wise federation and neighbor-based communication, as elaborated in the next section.

\section{Proposed Algorithm: L-FGADMM}
			\begin{algorithm}[t]
				\begin{algorithmic}[1]
				 \STATE {\bf Input}: \sm$N, f_n(\boldsymbol{\phi}_n)\;\forall n, \rho, K$\nm
				  \STATE {\bf Output}: \sm$\boldsymbol{\phi}_n\; \forall n$\nm 
				    \STATE {\bf Initialization}:  
				    \sm${\boldsymbol{\theta}_n^l}^{(0)}={\boldsymbol{\lambda}_n^l}^{(0)}=\boldsymbol{0}\; \forall n,l$\nm
    \WHILE{$k\leq K$}
					\STATE \hspace{-.3cm}\underline{Head worker $n \in {\cal N}_h$: in Parallel} 
					\STATE \hspace{.3cm}{Randomly selects} a mini-batch \sm$X_n^{(k)}$\nm

					\STATE \hspace{.3cm}{Updates} its primal variable \sm$\boldsymbol{\phi}_n^{(k+1)}$\nm via \eqref{headUpdate}
					\IF{$k$ mod $T_\ell=0$}
					\STATE \hspace{.3cm}{Transmits} \sm${\boldsymbol{\theta}_n^l}^{(k+1)}$\nm to its two tail neighbors
					\ENDIF
					\STATE \hspace{-.3cm}\underline{Tail worker $n \in {\cal N}_t$: in Parallel} 
					\STATE \hspace{.3cm}{Randomly selects} a mini-batch $X_n^{(k)}$
					\STATE \hspace{.3cm}{Updates} its primal variable \sm$\boldsymbol{\phi}_n^{(k+1)}$\nm via \eqref{tailUpdate}
					\IF{$k$ mod $T_\ell=0$}
					\STATE \hspace{.3cm}\text{Transmits} \sm${\boldsymbol{\theta}_n^l}^{(k+1)}$\nm to its two head neighbors
					\ENDIF

					\STATE \hspace{-.3cm}\underline{All workers: in Parallel}
					\IF{$k$ mod $T_\ell=0$}
					\STATE \hspace{.3cm}Updates the dual variables\!\! {\sm${\lamb_{n-1}^l}^{\!\!\!(k)}\!, {\lamb_{n}^l}^{(k)}$\nm}\!\!\! via~\eqref{lambdaUpdate}
					\ENDIF

\STATE $k \leftarrow k+1$
\ENDWHILE

   				\end{algorithmic}
				\caption{Layer-Wise Federated GADMM (L-FGADMM)} \label{L-FGADMM}				
			\end{algorithm}

To solve the problem defined in \eqref{com_admm}-\eqref{com_admm_c2}, in this section we propose L-FGADMM, by extending GADMM proposed in our prior work \cite{elgabli2019gadmm}. Following GADMM (see Fig.~\ref{fig1}b), workers in L-FGADMM are divided into head and tail groups, and communicate only with their neighboring workers. Compared to GADMM, L-FGADMM further improves communication efficiency through the following two ways. First, workers L-FGADMM periodically communicate as done in FL~\cite{Brendan17,samarakoon2018federated,KimCL:19}, in contrast to the communication per iteration in GADMM. Second, the communication period of L-FGADMM is adjusted separately for each layer (see Fig.~\ref{fig1}c), as opposed to GADMM and FL exchanging the entire models. L-FGADMM can thereby increase the communication periods for large-sized layers, while reducing the communication payload size.


To be specific, a physical network topology is converted into a logical worker connectivity graph in L-FGADMM. Then, the workers are split into a {\it head} group ${\cal N}_h$ and a {\it tail} group ${\cal N}_t$, such that each head worker is connected to neighboring tail workers. For the workers in the same group, their model parameters are updated in parallel, by iterating the mini-batch stochastic gradient descent algorithm (SGD). After $T_\ell$ iterations, the workers share the updated model parameters~$\boldsymbol{\theta}_n^l$ of the $\ell$-th layer with their neighbors. These operations~of L-FGADMM are summarized in Algorithm \ref{L-FGADMM}, and detailed next.


At first, the augmented Lagrangian of the problem in \eqref{com_admm}-\eqref{com_admm_c2} is defined as:

\vs\sm\begin{align}
\boldsymbol{\mathcal{L}}_{\rho}=&\sum_{n=1}^N f_n(\boldsymbol{\phi}_n, X_n^{(k)}) + \sum_{n=1}^{N-1}\sum_{l} {\lamb_n^l}^T (\boldsymbol{\theta}_{n}^l - \boldsymbol{\theta}_{n+1}^l)\nonumber\\ &+ \frac{\rho}{2}  \sum_{n=1}^{N-1}\sum_{l} \parallel \boldsymbol{\theta}_{n}^l - \boldsymbol{\theta}_{n+1}^l\parallel_2^2,
\label{augmentedLag4}
\end{align}\nm
where ${\lamb_n^l}$ is the $n$-th worker's dual variable of the $\ell$-th layer, and $\rho$ is a constant penalty term. For the sake of explanation, hereafter the $n$-th worker's model parameter vector $\boldsymbol{\theta}_{n}^l$ of the $\ell$-th layer is called a primal variable. Head and tail workers' primal and dual variables are updated through the following three phases.


\vspace{5pt}\noindent\textbf{1) Head primal updates.}\quad Head workers receive the primal variables from their tail workers, and update the dual variables at an interval of $T_\ell$ iterations. These variables at the $(k+1)$-th iteration are thus fixed as the values at the $\lfloor k/T_\ell\rfloor$-th iteration. Given these fixed primal and dual variables associated with neighbors, at iteration $k+1$, each head worker runs mini-batch SGD to minimize $\boldsymbol{\mathcal{L}}_{\rho}^{(k+1)}$. Applying the first-order condition to $\boldsymbol{\mathcal{L}}_{\rho}^{(k+1)}$ yields the $n$-th head worker's mode update as follows: 

\vs\sm\begin{align}
&{\boldsymbol{\phi}}_{n \in {\cal N}_h}^{(k+1)} =\underset{\phi_n}{\text{argmin}} \Big\{f_n(\boldsymbol{\phi}_n, X_n^{(k)})\nonumber\\&+\sum_{l}{\lamb_{n-1}^l}^T ({\boldsymbol{\theta}_{n-1}^l}^{(k)} - \boldsymbol{\theta}_{n}^l)+\sum_{l}{\lamb_n^{l}}^T (\boldsymbol{\theta}_{n}^l - {\boldsymbol{\theta}_{n+1}^l}^{(k)}) \nonumber\\ &+ \frac{\rho}{2} \sum_{l}\parallel {\boldsymbol{\theta}_{n-1}^l}^{(k)} - \boldsymbol{\theta}_{n}^l\parallel_2^2+ \frac{\rho}{2} \sum_{l}\parallel \boldsymbol{\theta}_{n}^l - {\boldsymbol{\theta}_{n+1}^l}^{(k)}\parallel_2^2\Big\}.
\label{headUpdate}
\end{align}\nm
After $T_\ell$ iterations, the head worker transmits the updated primal variable ${\boldsymbol{\theta}_n^\ell}^{(k+1)}$ to its two tail neighbors, workers $n-1$ and $n+1$.

\vspace{5pt}\noindent\textbf{2) Tail primal updates.}\quad 
Following the same principle in the head primal updates, at the $(k+1)$-th iteration, the $n$-th tail worker updates its model as:

\vs\sm\begin{align}
&\hspace{-7pt}{\boldsymbol{\phi}}_{n \in {\cal N}_t}^{(k+1)} =\underset{\phi_n}{\text{argmin}} \Big\{f_n(\boldsymbol{\phi}_n, X_n^{(k)})\nonumber\\
&\hspace{-7pt} +\sum_{l}{\lamb_{n-1}^l}^T ({\boldsymbol{\theta}_{n-1}^l}^{(k+1)} \!-\! \boldsymbol{\theta}_{n}^l) +\sum_{l}{\lamb_n^{l}}^T (\boldsymbol{\theta}_{n}^l - {\boldsymbol{\theta}_{n+1}^l}^{(k+1)}) \nonumber\\
&\hspace{-7pt}+ \frac{\rho}{2} \sum_{l}\parallel {\boldsymbol{\theta}_{n-1}^l}^{(k+1)} \!-\! \boldsymbol{\theta}_{n}^l\parallel_2^2 + \frac{\rho}{2} \sum_{l}\parallel \boldsymbol{\theta}_{n}^l - {\boldsymbol{\theta}_{n+1}^l}^{(k+1)}\parallel_2^2\Big\}.
\label{tailUpdate}
\end{align}\nm
After $T_\ell$ iterations, the tail worker transmits the updated primal variable ${\boldsymbol{\theta}_n^\ell}^{(k+1)}$ to its two head neighbors, workers $n-1$ and~$n+1$.

\vspace{5pt}\noindent\textbf{3) Dual updates.}\quad
After the updated tail primal variables are exchanged, every worker locally updates the dual variables $\lamb_{n-1}^l$ and $\lamb_n^l$ as follows: 

\vs\sm\begin{align}
{{\lamb}_n^l}^{(k+1)}={{\lamb}_n^l}^{(k)} + \rho({{\boldsymbol{\theta}}_{n}^l}^{(k+1)} - {{\boldsymbol{\theta}}_{n+1}^l}^{(k+1)}).
\label{lambdaUpdate}
\end{align}\nm

The convergence of the aforementioned primal and dual variable updates are theoretically proved for convex, proper, and smooth loss functions only when the entire layers are exchanged at every iteration~\cite{elgabli2019gadmm}. The convergence proof of L-FGADMM for different exchanging periods under deep NN architectures is deferred to future work. Meanwhile, the effectiveness of L-FGADMM is empirically corroborated in the next section.

\section{Numerical Evaluations}

\begin{table}[t]
\centering
\resizebox{\columnwidth}{!}{
\begin{tabular}{c c c } 
    \toprule
    \multicolumn{3}{c}{(a) \textbf{MLP} }\\
    Layer & Output Shape & \#Weights\\
    \midrule
    \textbf{\texttt{fc1}} & 256 & \textbf{200,960}\\
    \texttt{fc2} & 128 & 32,896\\
    \texttt{fc3} & 64 & 8,256\\
    \texttt{fc4} & 32 & 2,080\\
    \texttt{fc5} & 16 & 528\\
    \texttt{fc6} & 10 & 170\\
    \midrule
    Total & & 244,890\\   
    \bottomrule
\end{tabular}
\hspace{5pt}

\begin{tabular}{c c c } 
    \toprule
    \multicolumn{3}{c}{(b) \textbf{CNN} }\\
    Layer & Output Shape & \#Weights\\
    \midrule
    \texttt{conv1} & 28x28x8 & 208\\
    \texttt{conv2} & 14x14x6 & 3,216\\
    \texttt{conv3} & 7x7x32 & 8,224\\
    \textbf{\texttt{fc4}} & 400 & \textbf{627,600}\\
    \texttt{fc5} & 10 & 4,010\\\\
    \midrule
     Total & & 643,258\\   
    \bottomrule
\end{tabular}}
\caption{\small NN architectures: (a) MLP comprising 6 fully-connected layers (\texttt{fc1-6}) and (b) 5-layer CNN consisting of 3 convolutional layers (\texttt{conv1-3}) and 2 fully-connected layers (\texttt{fc4-5}).}\label{tab:1}
\end{table}

\begin{figure}[t]
    \centering
    \subfigure[MLP.]{\includegraphics[trim={0 .6cm 0 0}, clip, width=\columnwidth]{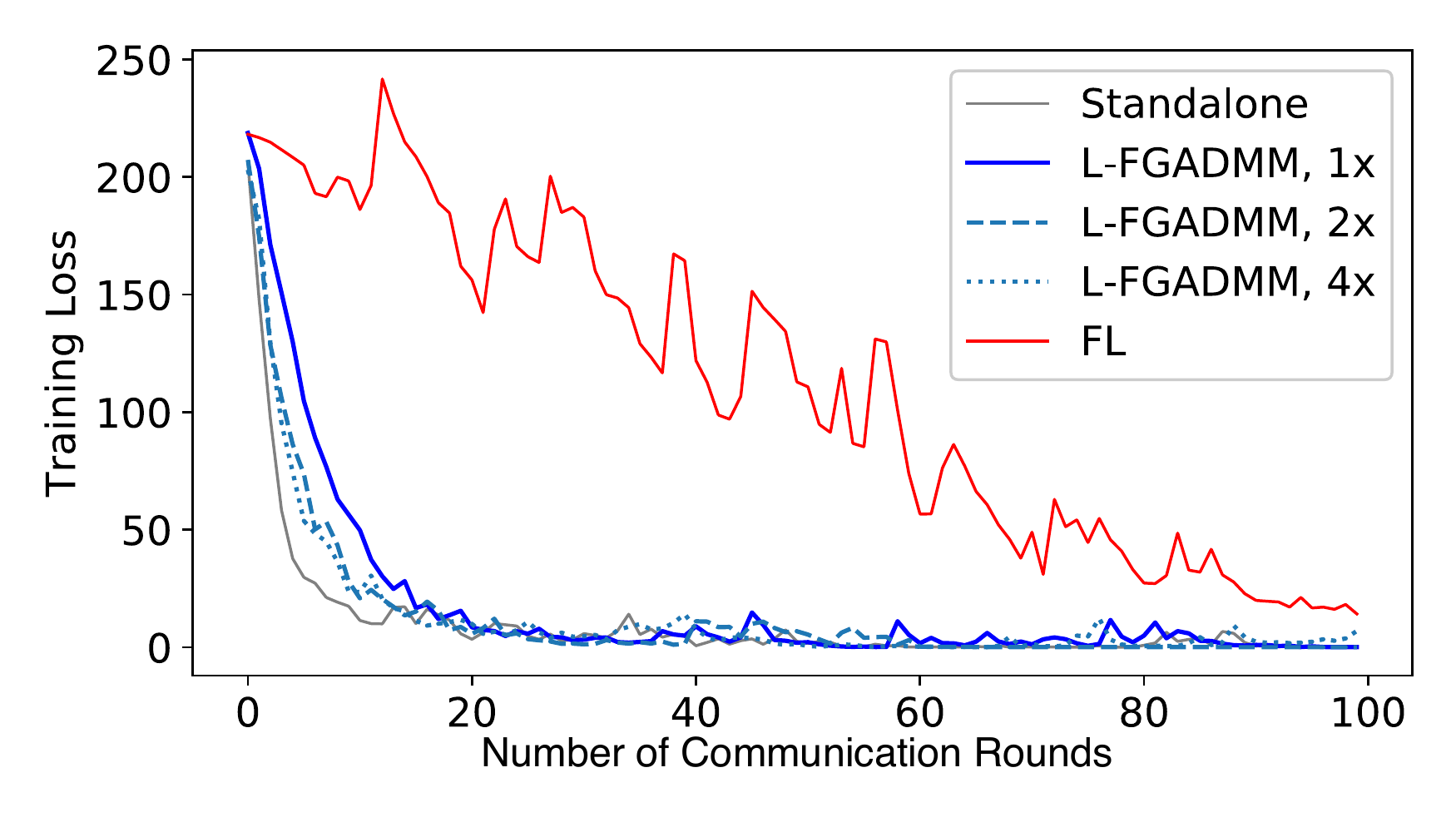}} 
    \subfigure[CNN.]{\includegraphics[trim={0 .6cm 0 0}, clip, width=\columnwidth]{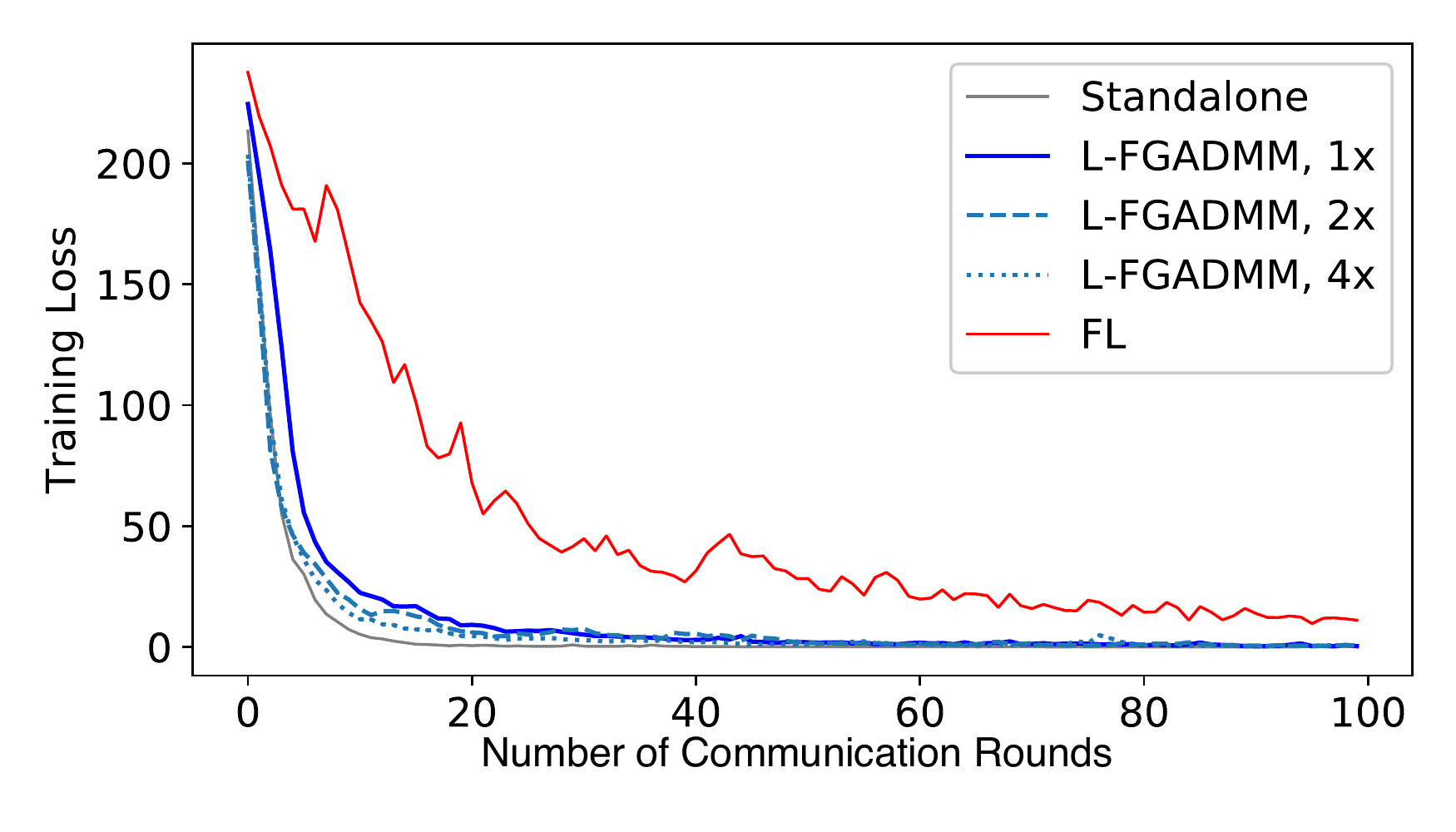}} 
    \caption{\small \emph{Training loss} of L-FGADMM under (a) MLP and (b) CNN, when the largest layer's exchanging period $T_{\ell_{\max} }$ is 2x or 4x longer.}
    \label{loss}
    \end{figure}

    \begin{figure}[t]
    \subfigure[MLP.]{\includegraphics[trim={0 .6cm 0 0}, clip, width=\columnwidth]{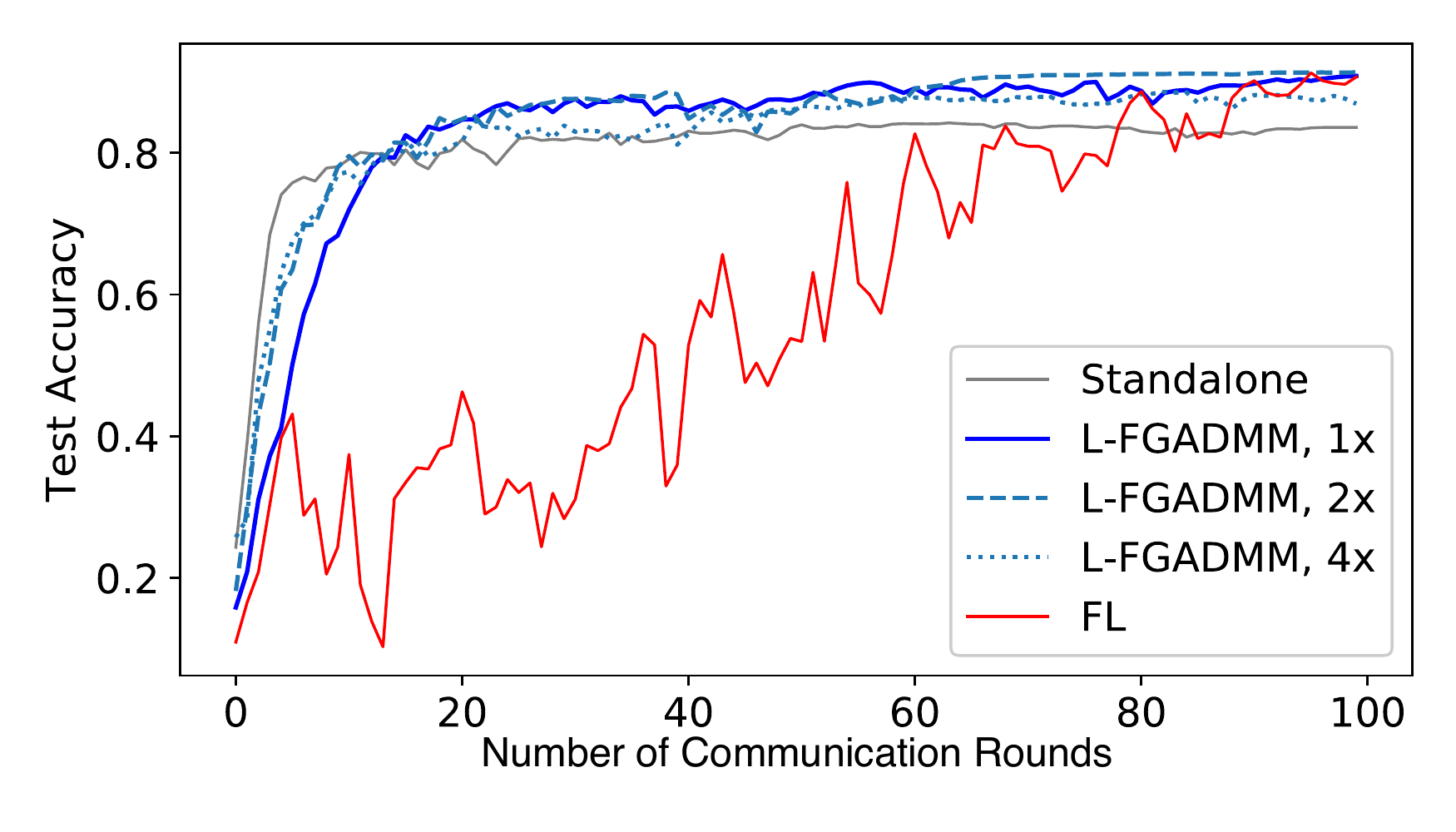}} 
    \subfigure[CNN.]{\includegraphics[trim={0 .6cm 0 0}, clip, width=\columnwidth]{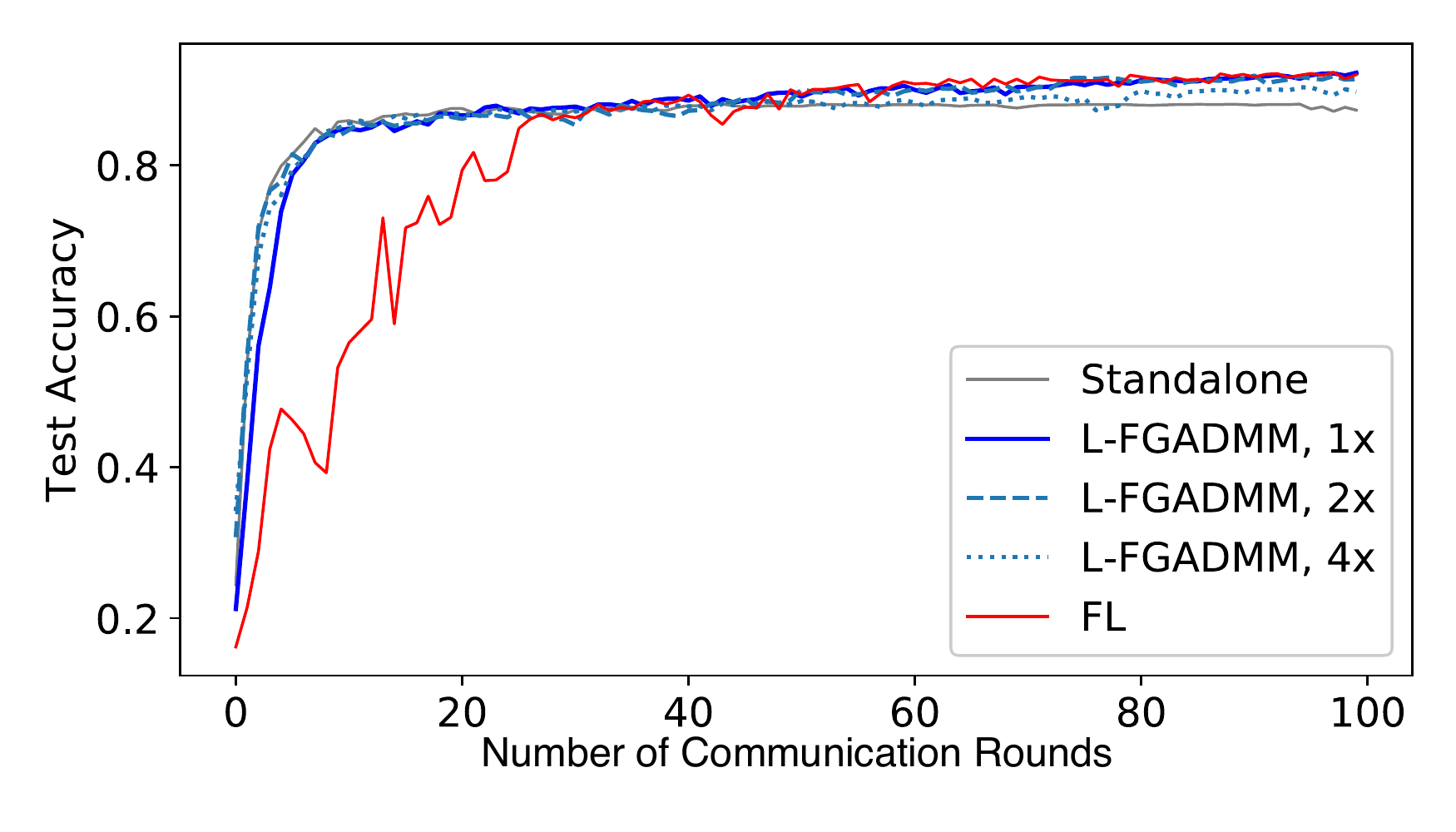}} 
    \caption{\small \emph{Test accuracy} of L-FGADMM under (a) MLP and (b) CNN, when the largest layer's exchanging period $T_{\ell_{\max} }$ is 2x or 4x longer.}
    \label{acc}
    \end{figure}

    \begin{figure*}[t]
        \centering
        \subfigure[MLP.]{\includegraphics[width=\columnwidth]{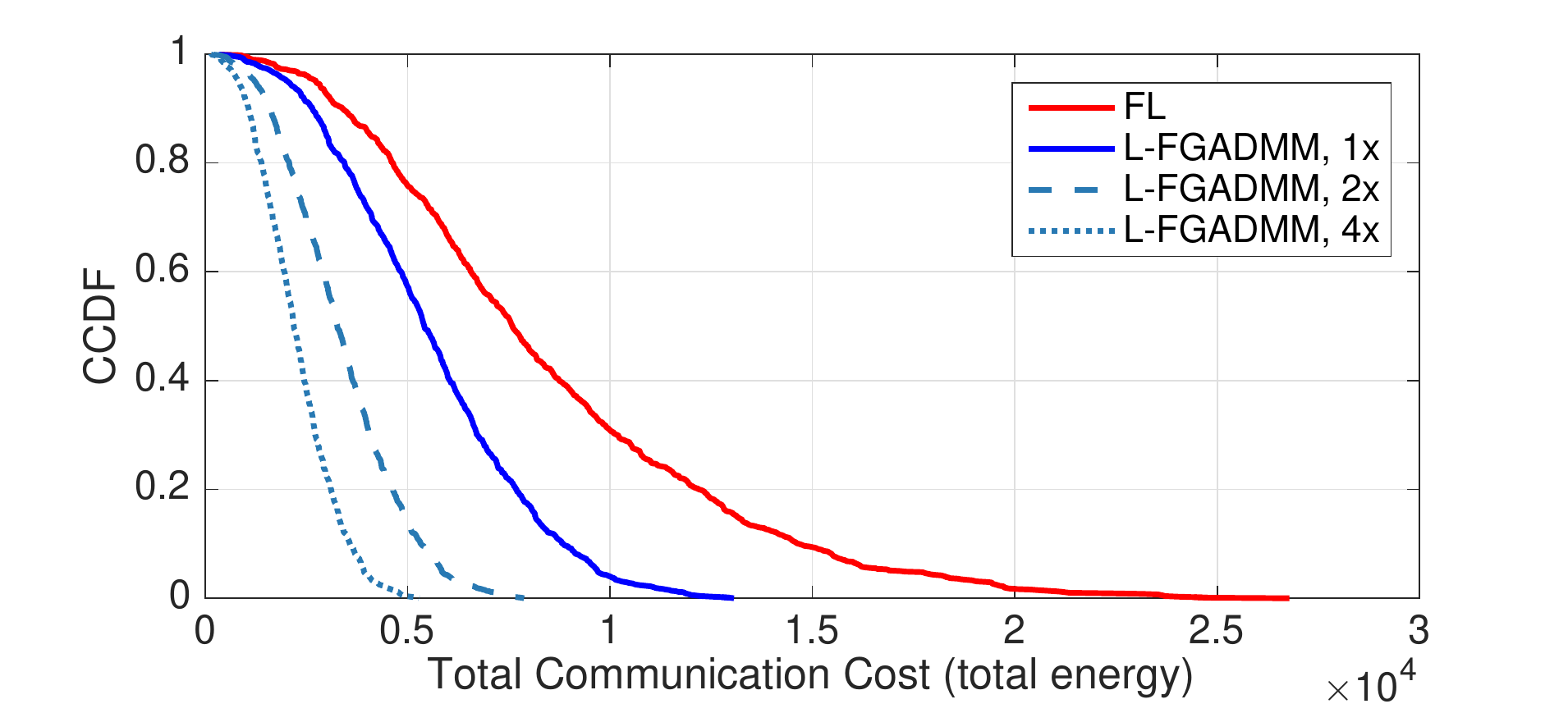}}
        \subfigure[CNN.]{\includegraphics[width=\columnwidth]{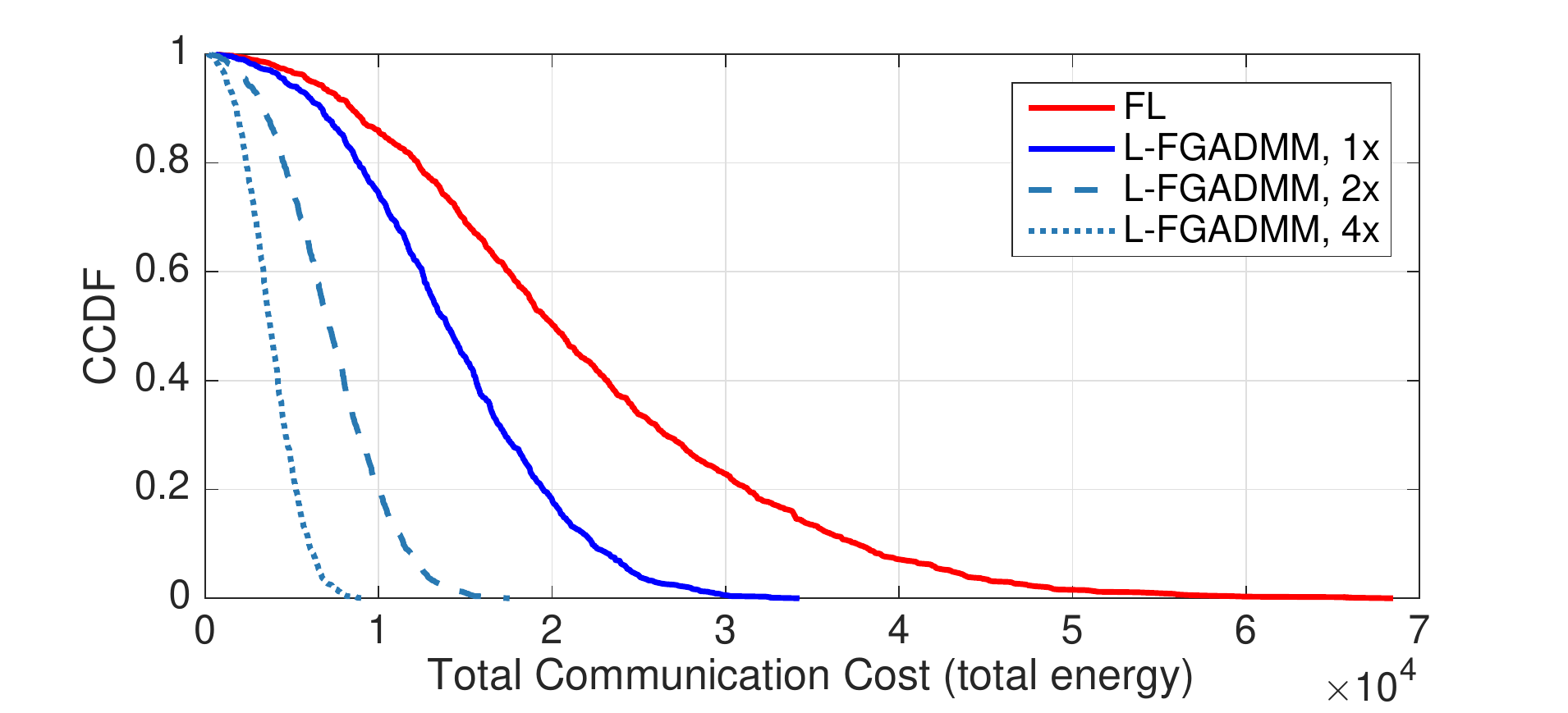}}
        \caption{\small \emph{Total communication cost} of L-FGADMM under (a) MLP and (b) CNN, when the largest layer's exchanging period $T_{\ell_{\max} }$ is 2x or 4x longer.}
        \label{cost}
        \end{figure*}

This section validates the performance of L-FGADMM for a classification task, with 4 workers uniformly randomly distributed over a 50x50 m$^2$ plane. These workers are assigned to head and tail groups, such that the length of the path starting from one worker passing through all workers is minimized. The simulation settings are elaborated as follows.

\vspace{3pt}\noindent\textbf{Datasets.}\quad
We consider the MNIST dataset comprising 28x28 pixel images that represent hand-written 0-9 digits. Each worker has $500$ training samples, independent and identically distributed across workers, and utilizes randomly selected $100$ samples per mini-batch SGD iteration, i.e., $X_n^{(k)}=100$.

\vspace{3pt}\noindent\textbf{Communication periods.}\quad
Denoting as $\ell_{\max}$ the largest layer, its communication period under L-FGADMM is set as $T_{\ell_{\max}}=\beta T_\ell$ for all $\ell\neq \ell_{\max}$. Hereafter, our proposed scheme is referred to as L-FGADMM 1x, 2x, or 4x, when $\beta=1$, $2$, or $4$, respectively. The communication periods of the other layers are identically set to 5 iterations.

\vspace{3pt}\noindent\textbf{NN architectures.}\quad
To examine the impact of NN architectures, two different NN models are considered, a \emph{multi-layer perceptron (MLP)} NN and \emph{convolutional neural network (CNN)}. As Table~\ref{tab:1} describes, the MLP consists of 6 layers, among which the 1st layer is the largest, i.e., $\ell_{\max}=1$, and has 82\% weight parameters of the entire model. The CNN comprises 5 layers, and the 4th layer is the largest among them, i.e., $\ell_{\max}=4$, having 97.6\% weight parameters of the entire model.

\vspace{3pt}\noindent\textbf{Baselines.}\quad
We compare L-FGADMM with two benchmark schemes, (i) \emph{FL} running mini-batch SGD with the learning rate $0.01$, while exchanging local gradients every 5 iterations~\cite{Brendan17}; and (ii) \emph{standalone} mini-batch SGD with a single worker. In FL, the worker having the minimum sum distances to all other workers is set as a parameter server. In the standalone case, there is no communication, but for the sake of convenience 5 SGD iterations are counted as a single communication round.

\vspace{3pt}\noindent\textbf{Performance measures.}\quad
The performance of each scheme is measured in terms of training loss, test accuracy, and total communication cost. \emph{Training loss} is measured using the cross entropy function. \emph{Test accuracy} is calculated as the fraction of correct classification outcomes. \emph{Total communication cost} is the sum of the total communication energy. With respect to these three figure of merits, the effectiveness of L-FGADMM is described as follows.


\begin{itemize}[leftmargin=10pt]
    \item \textbf{Fast convergence of L-FGADMM}: As shown by Fig.~\ref{loss}, under both MLP and CNN architectures, L-FGADMM 1x, 2x, and 4x converge within 100 communication rounds. For all cases, the final loss values of L-FGADMM are close to each other, which are up to 13.8\% smaller than FL whose training speed is also slower. It is noted that the standalone baseline yields the fastest convergence speed. This is because of its overfitting towards $500$ local training samples, which results in poor accuracy as explained next.

    \vspace{5pt}\item \textbf{High accuracy of L-FGADMM}: Fig.~\ref{acc}b shows that under CNN, L-FGADMM 1x achieves the highest final test accuracy (92.25\%), followed by FL (92\%), L-FGADMM 2x (91.42\%), L-FGADMM 4x (89.76\%), and the standalone case (87.3\%). It is noticeable that L-FGADMM 2x accuracy is comparable to FL, which converges faster while exchanging less layers than FL. Fig.~\ref{acc} demonstrates that under MLP, surprisingly, L-FGADMM 2x achieves the highest accuracy (91.37\%), followed by L-FGADMM 1x (90.87\%), FL (90.72\%), L-FGADMM 4x (86.94\%), and the standalone case (83.61\%). The excellence of L-FGADMM 2x can be explained by its regularization effect. Its skipping the largest layer communications introduces additional errors compared to L-FGADMM 1x, which exhibits better generalization.

    \vspace{5pt}\item \textbf{Low communication cost of L-FGADMM}: Fig.~\ref{cost} illustrates the complementary cumulative distribution function (CCDF) of the total communication cost (energy). We perform $1000$ experiments. Every run, we randomly drop a $4$ workers in $100$ x $100$ m$^2$ grid. We assume that the bandwidth per worker is $1$MHz ($B=1$MHz), the transmission power is $1$mw ($P=1$mw), and Noise spectral density is $1E-9$ ($N_0=1E-9$). We use the free space path loss model, so the SNR at the receiver $j$ when transmitter $i$ is transmitting is $P/(d_{i,j}^2N_0B)$, where $d_{i,j}$ is the distance between worker $i$ and $j$, and the achievable rate is computed using Shannon's formula. We assume every element in the model is transmitted using $32$ bits. Therefore, we compute the duration and the energy needed to transmit the shared model elements at every iteration, then we sum over all iterations and workers to find the total energy consumed by the system. The result shows that L-FGADMM achieves lower mean and variance of the total communication cost, compared to FL and the standalone baseline. Specifically, the mean total communication cost of L-FGADMM is up to 3.75x and 5.75x lower than FL, under MLP and CNN, respectively. Furthermore, the variance of L-FGADM is up to 22.5x and 51.4x lower than FL, under MLP and CNN, respectively. There are two rationales behind these results. In L-FGADMM, the worker connectivity is based on nearest neighbors in a decentralized setting, leading to shorter link distances than FL whose connectivity is centralized. Furthermore, the payload sizes are smaller thanks to partially skipping the largest layer exchanges, inducing higher communication efficiency of L-FGADMM.
        
\end{itemize}


\section{Conclusions}
By leveraging and extending GADMM and FL, in this article we proposed L-FGADMM, a communication-efficient decentralized ML algorithm that exchanges the largest layers of deep NN models less frequently than other layers. Numerical evaluations validated that L-FGADMM achieves fast convergence and high accuracy while significantly reducing the mean and variance of the communication cost. Generalizing this preliminary study, optimizing layer exchanging periods under different NN architectures and network topologies could be an interesting topic for future research.


\section*{Acknowledgement}
This research was supported in part by Academy of Finland (Grant Nr. 294128), in part by the 6Genesis Flagship (Grant Nr. 318927), in part by the Kvantum Institute Strategic Project (SAFARI), and in part by the Academy of Finland thorough the MISSION Project (Grant Nr. 319759).

\bibliographystyle{IEEEtran}
%


\end{document}